\documentclass[journal]{IEEEtran}

\IEEEoverridecommandlockouts
\usepackage{cite}

\ifCLASSINFOpdf
  \usepackage[pdftex]{graphicx}
\else
\fi
\usepackage{amsmath}
\usepackage{tikz}
\usepackage{textcomp}
\usepackage{hyperref}
\usepackage{lipsum}
\usepackage[]{algorithm2e}
\usepackage{pdfpages}

\newcommand\copyrighttext{%
	\footnotesize \textcopyright © 2019 IEEE. Personal use of this material is permitted. Permission from IEEE must be obtained for all other uses, in any current or future media, including reprinting/republishing this material for advertising or promotional purposes, creating new collective works, for resale or redistribution to servers or lists, or reuse of any copyrighted component of this work in other works.
}
		
\newcommand\copyrightnotice{%
	\begin{tikzpicture}[remember picture,overlay]
	\node[anchor=south,yshift=3pt] at (current page.south) {\fbox{\parbox{\dimexpr\textwidth-\fboxsep-\fboxrule\relax}{\copyrighttext}}};
	\end{tikzpicture}%
}

\begin{document}
	
%
\title{Enhancing approximation abilities of neural networks by training derivatives}

%
%
%

\author{V.I. Avrutskiy
\thanks{V.I. Avrutskiy is with the Department of Aeromechanics and Flight Engineering of Moscow Institute of Physics and Technology, Institutsky lane 9, Dolgoprudny, Moscow region, 141700, e-mail: avrutsky@phystech.edu}
}

%
%

\markboth{Submitted to IEEE Transactions on Neural Networks and Learning Systems}%
{Shell \MakeLowercase{\textit{et al.}}: Bare Demo of IEEEtran.cls for IEEE Journals}
%




\maketitle
\copyrightnotice
\begin{abstract}
	
A method to increase the precision of feedforward networks is proposed. It requires a prior knowledge of a target function derivatives of several orders and uses this information in gradient based training. Forward pass calculates not only the values of the output layer of a network but also their derivatives. The deviations of those derivatives from the target ones are used in an extended cost function and then backward pass calculates the gradient of the extended cost with respect to weights, which can then be used by any weights update algorithm. Despite a substantial increase in arithmetic operations per pattern (if compared to the conventional training), the extended cost allows to obtain 140--1000 times more accurate approximation for simple cases if the total number of operations is equal. This precision also happens to be out of reach for the regular cost function. The method fits well into the procedure of solving differential equations with neural networks. Unlike training a network to match some target mapping, which requires an explicit use of the target derivatives in the extended cost function, the cost function for solving a differential equation is based on the deviation of the equation's residual from zero and thus can be extended by differentiating the equation itself, which does not require any prior knowledge. Solving an equation with such a cost resulted in 13 times more accurate result and could be done with 3 times larger grid step. GPU-efficient algorithm for calculating the gradient of the extended cost function is proposed.
\end{abstract}


%
\IEEEpeerreviewmaketitle

\section{Introduction}
%
%
%
%
\IEEEPARstart{N}{eural} networks can be used as universal approximators\cite{hornik1989multilayer,cybenko1989approximation,kuurkova1992kolmogorov,hornik1991approximation} in a wide range of dimensions. In high-dimensional cases they successfully overcome the curse of dimensionality\cite{barron1993universal,barron1994approximation}, thus being an excellent remedy for problems like voice recognition\cite{deng2013machine,graves2006connectionist} and pattern classification\cite{bishop1995neural,krizhevsky2012imagenet,zhang2000neural}. Low-dimensional applications are not that famous since many alternatives are available. For example, in solving differential equations\cite{kumar2011multilayer} (usually 1D\cite{meade1994numerical,malek2006numerical} or 2D\cite{lagaris1998artificial}, sometimes 3D\cite{lagaris1997artificial}) neural networks inevitably have to compete with other methods like finite differences, where functions are described by their values on a set of points and in each point, in theory, those values can be accurate within machine precision. Such quality of approximation is not readily achievable for neural networks. Widely developed techniques for training them\cite{srivastava2014dropout,wan2013regularization,chang2015batch,goodfellow2013maxout,mishkin2015all} are mostly focused on problems like classification, which are not very sensitive to the actual output since small disturbances can hardly produce a shift in class. For the case of direct function approximation, any deviation of an output decreases the accuracy.

This paper proposes a method of utilizing information about target derivatives that increases the precision of neural networks. For some low-dimensional cases, it allows deviations from targets to come close to the rounding error of single precision used during the training, thus addressing the gap between describing a function by an array of values and by a neural network. The concept of using derivatives for approximation \cite{murray2005transformations} is quite common and was investigated for neural networks in numerous studies \cite{simard1998transformation,drucker1992improving,he2000multilayer,flake2000differentiating,cardaliaguet1992approximation,pukrittayakamee2011practical,basson1999approximation}, however, the implementations of training in said papers included only low order derivatives and used somewhat small architectures, since the conditions of tests did not lead to precision gains of few orders of magnitude. Even though requirements for architectures of neural networks to approximate derivatives are usually modest\cite{hornik1991approximation}, extra layers are sometimes necessary \cite{sontag1992feedback}.

Due to the necessity to train high order derivatives, which are not popular in applications nor implementations, this paper also includes an algorithm capable of an efficient high-order forward and backward procedures for arbitrary feedforward networks. It is derived directly from formulas for derivative transformation under a change of coordinates that are created by connections between layers, and wherever possible reductions are made. Somewhat similar algorithms can be found in papers on solving differential equations with neural networks\cite{lagaris1998artificial,he2000multilayer,berg2018unified}, however, no papers on this subject contain a universal algorithm for any order of derivatives and arbitrarily deep networks. Essentially the presented procedure is an equivalent to automatic differentiation \cite{griewank2008evaluating}, although software with similar capabilities is currently not quite optimized and fast research prototypes\cite{siskind2016efficient} are not yet available for GPUs. As the parallel computations are crucial for neural networks, all formulas are written in terms of matrix multiplications which are implemented on GPUs with highly efficient routines and element-wise operations which are parallelized most easily.
\section{Motivation}
A particular approach\cite{lagaris1998artificial,kumar2011multilayer} to solving differential equations was investigated: a single solution of one equation corresponds to one neural network which is treated as a smooth function. Its inputs are chosen as independent variables of the equation and the output is supposed to be the solution's value. A simplified procedure of obtaining such a network is as follows. At first, the weights are randomly initialized as they would be for the regular training \cite{glorot2010understanding,he2015delving}. Then, the network is substituted into the equation in place of the unknown function, which requires calculating the derivatives of the output with respect to the input. Since the network is not yet a solution, after substitution the residual exists. The next step is weights tuning. It requires finding the gradient of the residual with respect to the weights, which then can be used by a weights update algorithm. Two previous steps alternate each other until the residual becomes acceptable. The distinction from the conventional network training thus lies in the cost function which can contain the derivatives of the network.

Due to the lack of an algorithm that could handle cost functions with arbitrary derivatives in the previous papers on this subject, it was implemented and simple tests were conducted on a network with two inputs $x$, $y$, one output $z$ and a few hidden layers\footnote{the network and training conditions are described in the next section}. It was found that any derivative $D(z)$ of the output with respect to the inputs at least up to the 5\textsuperscript{th} order can be trained to approximate the derivative $D(f)$ of an analytical expression $f(x,y)$ using a cost function $e=[D(z)-D(f)]^{2}$ (here and further per-pattern cost functions $e$ will be used, the actual training cost $E$ is the sum of $e$ over all input patterns). The precision of approximation for this derivative did not depend on the order or the type of the derivative, and after 1000 epochs, the square root of the average cost was about $2.5\%$ of the standard deviation of $D(f)$. For the next test, a network was trained to fit the function and its derivatives simultaneously, using a cost $e=\sum c_{D}^{2}[D(z)-D(f)]^{2}$. The sum was running through all the 9 possible derivatives of order~$\leq3$ and the values themselves. Coefficients $c_{D}$ are the inversed standard deviations of the corresponding targets $D(f)$. Training with this type of cost function will be referred to as an extended one of the order 3. One could expect a lower or similar precision for each derivative and thus the root of the averaged cost to be $\sqrt{9+1}\cdot2.5\%\simeq8\%$. However, that was not the case. After 1000 epochs it reached $3.7\%$, thus making the deviation of each derivative lower. The gradient of the cost with respect to the weights increased roughly as the number of additional terms, but it still vanished with the same exponential rate as it was propagated backward. The particular precisions are presented in Table \ref{d3}: the root mean square of deviation for each derivative is measured in the percentage of the standard deviation of the corresponding target derivative and averaged along the same orders. The observed increase of precision for values (0\textsuperscript{th} derivative) led to a further investigation on how derivatives can be used to boost the precision most effectively.
\section{Results}
\begin{table}[!t]
	\renewcommand{\arraystretch}{1.5}
	\caption{The precision of various derivatives after 1000 epochs of the extended training with order $3$}
	\label{d3}
	\centering
	\begin{tabular}{c||c|c|c|c}
		\hline
		Derivative & 0 & 1 & 2 & 3 \\
		\hline
		$\overline{\text{rms}}$ & $0.07\%$ & $0.16\%$ & $0.63\%$ & $1.72\%$ \\
		\hline
	\end{tabular}
\end{table}
In all cases RProp\cite{riedmiller1993direct} is used for weights updating with parameters $\eta_{+}=1.2$, $\eta_{-}=0.5$. Weights are forced to stay in $[-20,20]$ interval, no min/max bonds for steps are imposed. Initial steps $\Delta_{0}$ are set to $2\cdot10^{-4}$ unless otherwise stated. When the number of epochs is greater than 5000, steps $\Delta$ that were reduced to zero are set back to $10^{-6}$ after each 8\% of epochs. Weights matrices are initialized\cite{glorot2010understanding,he2015delving} with random values from range $\pm2/\sqrt{||\kappa||}$, where $||\kappa||$ is the number of senders. Thresholds are from $\pm0.1$ range. All layers but the input and output are nonlinear with a sigmoid function
\begin{equation}\label{sigm}
\sigma(x)=\frac{1}{1+\exp(-x)}
\end{equation}
unless otherwise stated. All input patterns are processed in one batch. Root mean square values are always divided by the standard deviations of the corresponding functions.
\subsection{2D function approximation}
\begin{figure}[!t]
	\centering
	\includegraphics[width=3.6in]{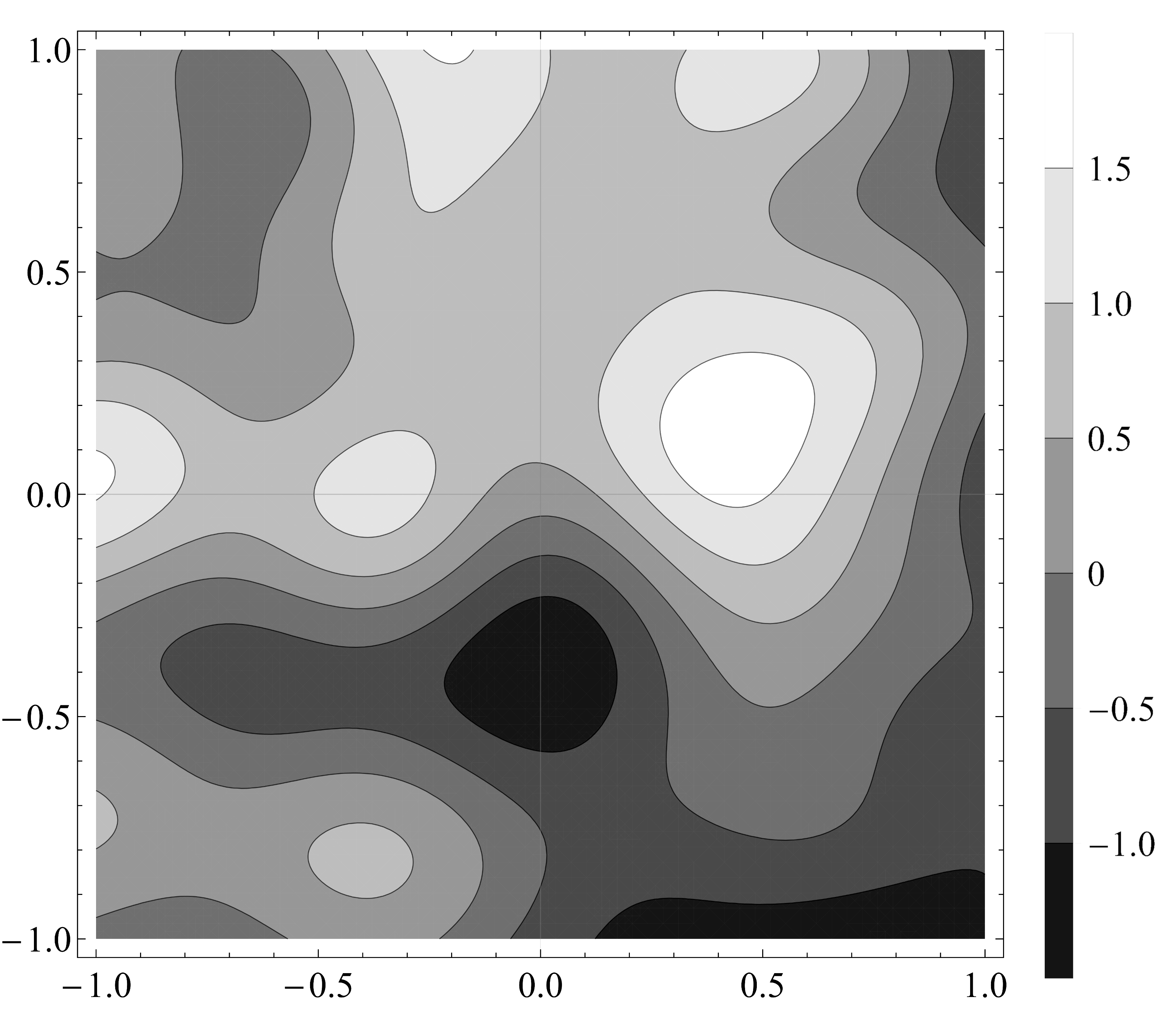}
	\caption{The function generated by a piece of random 2D Fourier series}
	\label{fig:2d}
\end{figure}

The target function is generated by a Fourier series:
\[
f(x,y)=\sum_{n=1}^{10}\sum_{k=1}^{10}\frac{r_{nk}}{n\cdot k}\sin nx\cos ky+\ldots
\]
Random coefficients $r_{nk}$ are uniformly distributed in $[-1,1]$. Three similar terms with other combinations of sine and cosine have separate coefficients and are omitted for brevity. The total number of parameters is 400. The region for approximation is a square $[-1,1]^{2}$. The particular realization is shown on Fig. \ref{fig:2d}. The input is generated as the vertices of a Cartesian grid with the outer points lying on the boundary, and has 729 points unless otherwise stated. In each point all derivatives are calculated analytically. After training the performance is measured on a grid with 9025 points. The network is a fully connected perceptron with the following configuration of layers (asterisks denote linear activation functions): $2^{*},128,128,128,128,1^{*}$.

\begin{table}[!t]
	\renewcommand{\arraystretch}{1.5}
	\caption{The precision of values after 1000 epochs of the extended training of order $d$}
	\label{vard}
	\centering
	\begin{tabular}{c||c|c|c|c|c|c}
		\hline
		$d$ & 0 & 1 & 2 & 3 & 4 & 5 \\
		\hline
		$\text{rms}$ & $2.5\%$ & $0.8\%$ & $0.09\%$ & $0.07\%$ & $0.28\%$ & $0.5\%$\\
		\hline
	\end{tabular}
\end{table}
Table \ref{vard} was created as an attempt to determine the optimal order of the extended training. It varies the maximum order of derivatives used in the cost $e=\sum c_{D}^{2}[D(z)-D(f)]^{2}$, the training was run for 1000 epochs. The first and the second derivatives boost the precision of values, however, when the order becomes higher than 3, it starts to decrease. Even inclusion of the third derivatives becomes ineffective if one takes into account the additional computational burden. Consider the 4\textsuperscript{th} order training: the cost contains 15 terms, only one of which is related to the precision of values. Moreover, the relative magnitude of those terms is increasing with the order, thus, it is understandable that the training is less focused on the zero order term. By choosing smaller $c_{D}$ coefficients for the 4\textsuperscript{th} order terms, it is possible to increase the precision of values, but no greater than it was for $d=3$. However, it was found that if during the training $c_{D}$ for higher orders are abruptly set to zero, the precisions of lower orders quickly increase. With further experiments, the following process was constructed: the extended training is started with an order $d$. After a certain number of epochs, it is terminated and re-initialized with the order $d-1$ and so on, up until only values alone ($d=0$) are trained. An equal number of epochs was chosen for all $d+1$ steps of this procedure. The training set remains the same throughout the process. This procedure will be referred to as an exclusion training of the order $d$. The re-initialization of RProp after turning off higher orders needs to be made with $\Delta=10^{-5}$, otherwise networks are disturbed too much. A gradual decrease of $c_{D}$ instead of an abrupt drop worked a bit worse. Inclusion of new derivatives, on the contrary, leads to the opposite effect: the precision degrades heavily until new derivatives are properly trained, and then slowly reaches its typical value for the new order.

\begin{figure}[!t]
	\centering
	\includegraphics[width=3.4in]{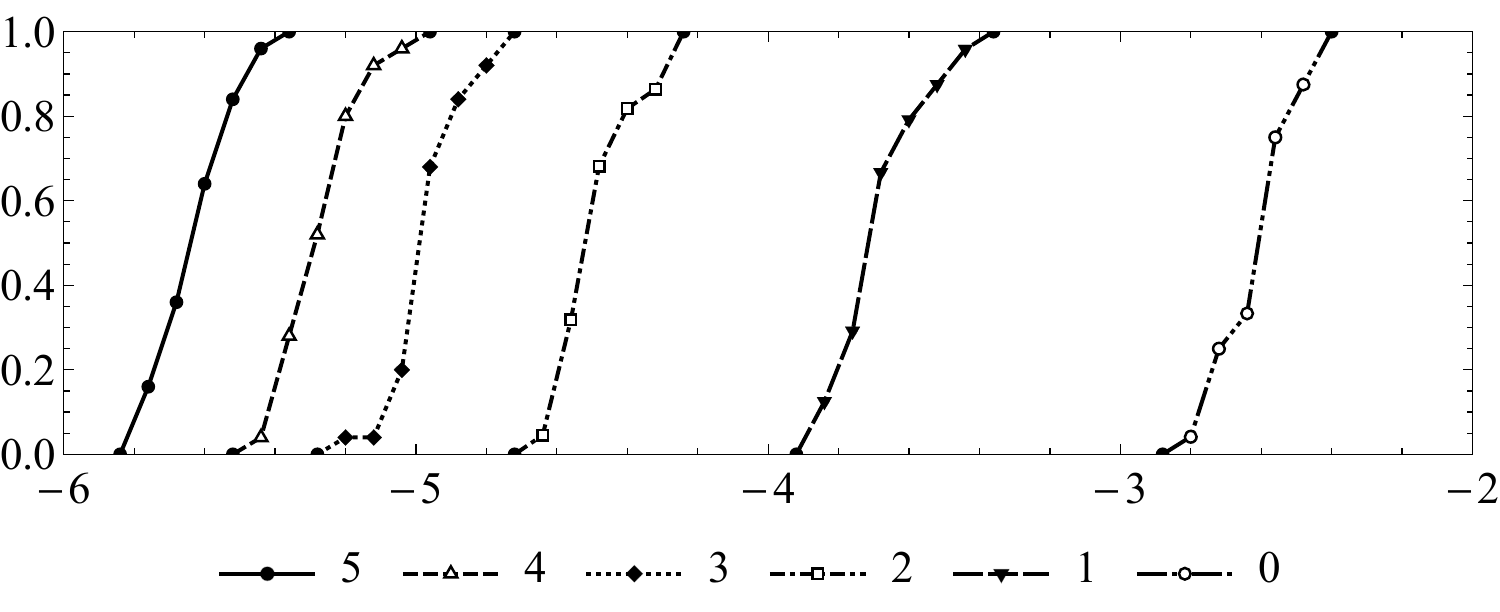}
	\caption{2D Function approximation: cumulative distribution functions for $\log_{10}(\text{rms}(z-f(x,y)))$ for different orders of the exclusion training. The best result of a lower order barely reaches the worst result of a higher order.}
	\label{cdf}
\end{figure}

Table \ref{final} summarizes the results obtained by various orders of the exclusion training. The median (denoted by tilde) of the root mean square of deviations for values $z-f(x,y)$ is taken across 25 training attempts for each order $d$. Distribution functions for precisions can be seen on fig. \ref{cdf}. Durations in (kilo) epochs are chosen to equalize the total number of arithmetic operations with the base level taken as 1750 epochs for each step of the order 5 on a grid with 729 points. Due to an overfitting encountered for lower orders, the number of points had to be increased, however, the least favorable scenario for higher order training was chosen: the number of samples was compensated to prevent overfitting, but the number of epochs (now marked by an asterisk) was chosen as if the number of samples remained the same. As soon as the overfitting is stopped, a further increase in number of points does not affect the precision. Training with $d=4$ and $d=5$ demonstrated no overfitting with 729 patterns even though the later steps of the training completely discarded high-order terms (provided the weights were not disturbed too much, $\Delta=10^{-5}$). The complexity for various orders is discussed in section V.
\begin{table}[!t]
	\renewcommand{\arraystretch}{1.3}
	\caption{2D Function approximation: the precision of the exclusion training of order $d$, the gain over the one of $d-1$. The number of kilo epochs (KE) equalizes the total amount of operations}
	\label{final}
	\centering
	\small
	\setlength\tabcolsep{1pt}
	\begin{tabular}{c||c|c|c|c|c|c}
		\hline
		$d$ & 0 & 1 & 2 & 3 & 4 & 5 \\
		\hline\hline
		$\widetilde{\text{rms}}$ & $2.3\cdot10^{-3}$ & $2\cdot10^{-4}$ & $3\cdot10^{-5}$ & $1\cdot10^{-5}$ & $5.2\cdot10^{-6}$ & $2.2\cdot10^{-6}$\\
		\hline
		gain & -- & 11.5 & 6.6 & 3 & 1.9 & 2.4\\
		\hline
		KE & 107\textsuperscript{*} & 26.5\textsuperscript{*} & 10.5\textsuperscript{*} & 5.2\textsuperscript{*} & 2.9 & 1.75\\
		\hline
	\end{tabular}
\end{table}
\subsection{Autoencoder for 3D curve}
Targets are the points of 3D space located on 1D curve:
\begin{equation}\label{curve}
(\cos t,\sin t,t/\pi)\equiv(f_{1},f_{2},f_{3}).
\end{equation}
It is possible to reconstruct their 1D representation using a network with 1-neuron linear layer\cite{petridis2016deep} inserted in the middle: $3^{*},64,64,1^{*},64,64,3^{*}$, provided it is trained to replicate the input (other layers are smaller as this task is less demanding). This network can be used to generate new points on the curve\cite{bourlard1988auto,bengio2009learning}. The training and test sets are generated by formula \eqref{curve} with 64 and 1184 equidistant values of $t\in[-2\pi,2\pi]$ respectively unless otherwise stated. The derivatives of $(f_{1},f_{2},f_{3})$ with respect to $t$ are calculated analytically. The network's output is a vector $z_{i}$ so the cost includes two sums:
\[
e=\sum_{D,i}c_{D,i}^{2}[D(z_{i})-D(f_{i})]^{2}=\sum_{j=1}^{d}\sum_{i=1}^{3}c_{ji}^{2}[\frac{\partial^{j}}{\partial t^{j}}z_{i}-\frac{\partial^{j}}{\partial t^{j}}f_{i}]^{2}.
\]
\begin{table}[!t]
	\renewcommand{\arraystretch}{1.3}
	\caption{3D Autoencoder: the precision of the exclusion training of order $d$, the gain over the one of $d-1$. The number of kilo epochs (KE) equalizes the total amount of operations}
	\label{finalspiral}
	\centering
	\small
	\setlength\tabcolsep{1pt}
	\begin{tabular}{c||c|c|c|c|c|c}
		\hline
		$d$ & 0 & 1 & 2 & 3 & 4 & 5 \\
		\hline\hline
		$\widetilde{\text{rms}}$ & $6.6\cdot10^{-4}$ & $4.1\cdot10^{-4}$ & $3.7\cdot10^{-5}$ & $2.2\cdot10^{-5}$ & $6.6\cdot10^{-6}$ & $4.7\cdot10^{-6}$\\
		\hline
		gain & -- & 1.5 & 11 & 1.7 & 3.4 & 1.3 \\
		\hline
		KE & 29\textsuperscript{*} & 9.5\textsuperscript{*} & 4.6\textsuperscript{*} & 2.6\textsuperscript{*} & 1.7 & 1.2\\
		\hline
	\end{tabular}
\end{table}
Table \ref{finalspiral} summarizes the results. The median (denoted by tilde) of the root mean square of deviations for the values of the first component: $z_{1}-f_{1}$ is taken across 25 training attempts for each order $d$. Two other components have similar precisions and gains. Durations in (kilo) epochs equalize the total number of arithmetic operations with the base level taken as 1200 epochs for each step of the order 5 exclusion training on 64 input vectors. The overfitting for orders $\le3$ is compensated by increasing the number of vectors to 128, but similarly to the previous case the number of epochs was not lowered. The complexity is discussed in section V.
\subsection{Solving differential equations}
Previous examples require a prior calculation of the target derivatives in order to increase the precision. Usually this kind of information is not obtained easily, especially for real world applications. However, for solving differential equations this technique can be used without any extra information. The minimizing procedure is based on the equation itself which can be differentiated as many times as necessary without any prior knowledge about the solution.

Consider a generic boundary value problem inside a 2D region $\Gamma$ with a boundary $\partial\Gamma$ for a function $u(x,y)$:
\[U(x,y,u,u_{x},u_{y},...)=0,\left.u\right|_{\partial\Gamma}=f.\]
There are different approaches, but the most accurate results are obtained using the method from \cite{lagaris1998artificial}. A new function $v$ is introduced with relation 
\begin{equation}\label{podst}
u(x,y)=v(x,y)\cdot\phi(x,y)+f.
\end{equation}
Here $\phi$ is a smooth function carefully chosen to vanish on $\partial\Gamma$: 
\[\left.\phi\right|_{\partial\Gamma}=0\]
and not to vanish anywhere inside the region. Usually, it is the simplest analytical expression that is zero on the boundary and has the maximum value of the order of 1 somewhere inside $\Gamma$. For trivial boundaries like the circle of radius $1$, one can choose $\phi=1-x^2-y^2$.
After the substitution, the equation is written for $v$:
\[V(x,y,v,v_{x},v_{y},...)=0.\]
And the boundary condition is
\[
\left.v\right|_{\partial\Gamma}<\infty.
\] 
The function $v$ can now be approximated with a neural network with two inputs and one output. As long as its values do not diverge during the training the boundary condition is satisfied. A discrete grid in the region $\Gamma$ is created and according to \cite{lagaris1998artificial} a cost function $e=V^{2}$ can be minimized with respect to the weights for all the grid points. Here, however, an additional derivatives of $V$ up to a certain order are included in the cost
\[
e=\sum[D(V)]^{2}.
\]
For example, up to the second order:
\[
V^{2}+\left(\frac{\partial V}{\partial x}\right)^{2}+\left(\frac{\partial V}{\partial y}\right)^{2}+\left(\frac{\partial^{2}V}{\partial x^{2}}\right)^{2}+\left(\frac{\partial^{2}V}{\partial x\partial y}\right)^{2}+\left(\frac{\partial^{2}V}{\partial y^{2}}\right)^{2}.
\]

The method was tested on a boundary value problem inside a circle $\Gamma:x^2+y^2\le1$ for Poisson's equation with a nonlinear source
\[
u_{xx}+u_{yy}=u^{2}+\frac{3}{2}u^{3},\left.u\right|_{\partial\Gamma}=-2.
\]
Numerical results are compared against the analytical solution:
\[
u_{\text{a}}=\frac{4}{x^{2}+y^{2}-3}.
\]
A Cartesian grid inside $[-1,1]^{2}$ is created with spacing $\lambda$. All points outside $\Gamma$ are excluded and additional points from the boundary with spacing $\lambda$ are added. This task is similar to the one from the subsection A, however, it is simpler and does not require as many hidden neurons. For differential equations deeper networks seem to perform better. The configuration was chosen as $2^{*},64,64,64,64,64,64,1^{*}$.
In this case, instead of giving low-order training extra advantages whenever it encounters overfitting, the total number of arithmetics is to be equalized and the increase of sample points will be accounted for. Since this results in a trade-off between the number of epochs and the size of a grid, it was verified that in all cases using grids that produce overfitting in order to increase the number of epochs decreased the quality of the solution.
\begin{table}[!t]
	\renewcommand{\arraystretch}{1.3}
	\caption{Poisson's equation: the rms of residual $V$ for the exclusion training of order $d$, the gain over the one of $d-1$, the rms of deviation from the analytical solution $u_{\text{a}}$. The number of kilo epochs (KE) equalizes the total amount of operations}
	\label{finalpde}
	\centering
	\small
	\setlength\tabcolsep{1pt}
	\begin{tabular}{c||c|c|c|c}
		\hline
		$d$ & 0 & 1 & 2 & 3 \\
		\hline\hline
		$\widetilde{\text{rms}}(V)$ & $3\cdot10^{-3}$ & $5.4\cdot10^{-4}$ & $1.8\cdot10^{-4}$ & $9.3\cdot10^{-5}$\\
		\hline
		gain & -- & 5.5 & 3 & 2 \\
		\hline
		$\widetilde{\text{rms}}(u-u_{\text{a}})$ & $2.8\cdot10^{-5}$ & $9\cdot10^{-6}$ & $3.6\cdot10^{-6}$ & $2.1\cdot10^{-6}$ \\
		\hline
		$\lambda$ & 0.052 & 0.1 & 0.125 & 0.15\\
		\hline
		grid size & 1210 & 352 & 233 & 157\\
		\hline
		KE & 1.4 & 1.6 & 1.2 & 1\\
		\hline
	\end{tabular}
\end{table}
Networks are verified against the analytical solution on a Cartesian grid with about 8000 points. The results are presented in Table \ref{finalpde}. One can see that the deviation from analytical solution for $d=3$ comes somewhat close to its minimum possible value determined by the rounding error. If it was the only source, rms would be around $0.36\cdot10^{-6}=10^{-6.44}$. The complexity for various orders is discussed in section V.
\section{Algorithm}
This section is focused on implementing a gradient algorithm that can be used to minimize a more general form of a cost function for a neural network. The named function now depends not only on the network's output, but also on the derivatives of the output with respect to the input. It is valid for feedforward networks with any number of hidden layers and the derivatives of any order. Presented without thorough derivation, which can be found in the author's paper \cite{avrutskiy2017backpropagation}.
\subsection{Notation}
The quantity that each neuron obtains from the previous layer before applying
its nonlinear sigmoid mapping will be referred to as the neuron activity.
All neuron activities of a layer are gathered in a matrix with
the letter $z$ like $z^{\kappa\alpha}$. The index $\alpha$ runs through
the number of patterns and $\kappa$ through the neurons of that layer. The norm of indices such as $||\alpha||$ and $||\kappa||$ denotes 
the total number of enumerated objects. Here it refers to the number of input patterns and the number of neurons in the layer respectively. Different layers are denoted by different Greek letters which also appear in
weights matrices connecting those layers: $W^{\theta\kappa}$ would
be a matrix that is used to pass from the layer with neurons denoted by
$\kappa$ to the layer with those denoted by $\theta$. The input layer is denoted by $\beta$ and the output layer by $\omega$. In addition to
neurons activities $z^{\kappa\alpha}$, each layer has the derivatives
of those activities with respect to certain variables. Those derivatives are matrices
of the same size $||\kappa||\times||\alpha||$. When variables are known explicitly, they are denoted
by Latin letters $a,b,c,d$. The corresponding derivatives of $z^{\kappa\alpha}$ are denoted by lower indices:
\[
\frac{\partial}{\partial a}z^{\kappa\alpha}=z_{a}^{\kappa\alpha},
\frac{\partial^{2}}{\partial a^{2}}z^{\kappa\alpha}=z_{aa}^{\kappa\alpha}
\]
and so on. For formulas that allow an arbitrary number of variables, multi-index notation\cite{saint2018elementary} is used. Namely, if the order of variables is fixed: $(a,b,c,d,\dots)$, any derivative can be written using a multi-index $s=(s_{1},s_{2},\dots)$, which is a tuple of integers each component of which is the order of the derivative with respect to the corresponding variable. Thus,
\[\partial^{s}\equiv\frac{\partial^{|s|}}{\partial a^{s_{1}}\partial b^{s_{2}}\ldots}.\]
The absolute value of a multi-index is the sum of its components $|s|=\underset{i}{\sum}s_{i}$. The difference of two multi-indices is defined element-wisely.
\subsection{Forward pass}
The first step is to initialize $z^{\beta\alpha}$, $z_{a}^{\beta\alpha}$,
$z_{aa}^{\beta\alpha}$, $z_{b}^{\beta\alpha}$, $\ldots$ at the
input layer of the network. $z^{\beta\alpha}$ is simply the matrix of $||\alpha||$ input vectors of length $||\beta||$. Let us say $a$ is the first component of
the input, then for any $\alpha$: $z_{a}=(1,0,0,\dots)$ and the input matrix
is $z_{a}^{\beta\alpha}=\delta_{\beta-1}$ (the Kronecker delta). All higher derivatives
that include $a$ like $z_{ab}^{\beta\alpha}$, $z_{aa}^{\beta\alpha}$
are zero. Therefore, initializing derivatives with respect to the input
of a neural network is trivial. If $b$ is a variable that all components
of the input vector depend on, for example, let $||\beta||=2$ and $\alpha$ be fixed: $z=(\sin b,\cos b)$,
then $z_{b}=(\cos b,-\sin b)$. If $b$ depends on $\alpha$ (see section III.B) this derivative varies for other input vectors. Higher derivatives are non zero and should be calculated accordingly.

After the input layer matrix is generated and its derivatives are calculated,
they are to be propagated forward. Consider two generic successive
layers with indices $\kappa$ and $\theta$. The values themselves are propagated by a nonlinear sigmoid and a matrix multiplication (denoted by $\times$):
\begin{equation}\label{zeroth}
z^{\theta\alpha}=t^{\theta}+W^{\theta\kappa}\times\sigma(z^{\kappa\alpha}).
\end{equation}
Here $t^{\theta}$ is a threshold vector added to each column of the matrix
product to the right of it. The scalar function $\sigma$ is applied
independently to each element of $z^{\kappa\alpha}$. Due to linearity of \eqref{zeroth} and weights with thresholds being constant, one can apply a differential operator to both sides to establish the rule for derivative propagation from $\kappa$ to $\theta$:
\begin{equation}\label{fpassformula}
\partial^{s}z^{\theta\alpha}=W^{\theta\kappa}\times\left[\partial^{s}\sigma(z^{\kappa\alpha})\right].
\end{equation}
The term $\partial^{s}\sigma(z^{\kappa\alpha})$ is to be obtained using the chain rule. For example, the first order derivative with respect to $a$ is
\begin{equation}\label{za}
z_{a}^{\theta\alpha}=W^{\theta\kappa}\times\frac{\partial}{\partial a}\sigma(z^{\kappa\alpha})=W^{\theta\kappa}\times\left[\sigma'(z^{\kappa\alpha})\cdot z_{a}^{\kappa\alpha}\right].
\end{equation}
The expression in square brackets is an element-wise product. In similar
formulas the sigmoid argument as well as ``$\cdot$'' sign will be
omitted, so it is written as $\sigma'z_{a}^{\kappa\alpha}$.
The chain rule expresses the derivative of the next layer $z_{a}^{\theta\alpha}$ in terms of the derivative of the previous layer $z_{a}^{\kappa\alpha}$. The second order derivative propagation:
\begin{align}\label{mixed}
z_{ab}^{\theta\alpha} &= W^{\theta\kappa}\times\frac{\partial^{2}}{\partial a\partial b}\sigma(z^{\kappa\alpha})=\nonumber\\
 &= W^{\theta\kappa}\times\left[\sigma''z_{a}^{\kappa\alpha}z_{b}^{\kappa\alpha}+\sigma'z_{ab}^{\kappa\alpha}\right].
\end{align}
For non-mixed second derivatives like $\partial^{2}/\partial a^{2}$,
one can simplify the square brackets to $\sigma''\left[z_{a}^{\kappa\alpha}\right]^{2}+\sigma'z_{aa}^{\kappa\alpha}$ (the second power of square brackets is also an element-wise operation).
The same holds for further formulas: all Latin variables are considered
distinct, and in case they are not, it is useful to simplify expressions
first to avoid unnecessary arithmetic and/or memory operations. Any
mixed second derivative depends on three terms of a previous layer:
said second derivative plus both first order ones. Non-mixed second
derivatives depend only on two terms: the first and the second derivatives
with respect to that variable.
The third order:
\begin{align}\label{3rd}
z_{abc}^{\theta\alpha}&= W^{\theta\kappa}\times\Bigl[\sigma'''z_{a}^{\kappa\alpha}z_{b}^{\kappa\alpha}z_{c}^{\kappa\alpha}\nonumber\\
& +\sigma''\cdot(z_{ab}^{\kappa\alpha}z_{c}^{\kappa\alpha}+z_{ac}^{\kappa\alpha}z_{b}^{\kappa\alpha}+z_{bc}^{\kappa\alpha}z_{a}^{\kappa\alpha})+\sigma'z_{abc}^{\kappa\alpha}\Bigr].
\end{align}
The forth order:
\begin{align}\label{4th}
z_{abcd}^{\theta\alpha}=& W^{\theta\kappa}\times\Bigl[\sigma^{IV}z_{a}^{\kappa\alpha}z_{b}^{\kappa\alpha}z_{c}^{\kappa\alpha}z_{d}^{\kappa\alpha}\nonumber\\
& +\sigma'''(z_{ab}^{\kappa\alpha}z_{c}^{\kappa\alpha}z_{d}^{\kappa\alpha}+z_{ac}^{\kappa\alpha}z_{b}^{\kappa\alpha}z_{d}^{\kappa\alpha}+z_{ad}^{\kappa\alpha}z_{b}^{\kappa\alpha}z_{c}^{\kappa\alpha}\nonumber\\
& +z_{bc}^{\kappa\alpha}z_{a}^{\kappa\alpha}z_{d}^{\kappa\alpha}+z_{bd}^{\kappa\alpha}z_{a}^{\kappa\alpha}z_{c}^{\kappa\alpha}+z_{cd}^{\kappa\alpha}z_{a}^{\kappa\alpha}z_{b}^{\kappa\alpha})\nonumber\\
& +\sigma''(z_{a}^{\kappa\alpha}z_{bcd}^{\kappa\alpha}+z_{b}^{\kappa\alpha}z_{acd}^{\kappa\alpha}+z_{c}^{\kappa\alpha}z_{abd}^{\kappa\alpha}+z_{d}^{\kappa\alpha}z_{abc}^{\kappa\alpha}\nonumber\\
& +z_{ab}^{\kappa\alpha}z_{cd}^{\kappa\alpha}+z_{ac}^{\kappa\alpha}z_{bd}^{\kappa\alpha}+z_{ad}^{\kappa\alpha}z_{bc}^{\kappa\alpha})+\sigma'z_{abcd}^{\kappa\alpha}\Bigr].
\end{align}
The fifth order simplified for the cases considered in this paper:
\begin{align}\label{fifth}
z_{aaaaa}^{\theta\alpha}=&
W^{\theta\kappa}\times\Bigl[\sigma^{V}[z^{\kappa\alpha}_{a}]^{5}+\sigma^{IV}10[z^{\kappa\alpha}_{a}]^{3}z^{\kappa\alpha}_{aa}\nonumber\\
&
+\sigma'''(15z^{\kappa\alpha}_{a}[z^{\kappa\alpha}_{aa}]^{2}+10[z^{\kappa\alpha}_{a}]^{2}z^{\kappa\alpha}_{aaa})\nonumber\\
&
+\sigma''(10z^{\kappa\alpha}_{aa}z^{\kappa\alpha}_{aaa}+5z^{\kappa\alpha}_{a}z^{\kappa\alpha}_{aaaa})+\sigma'z^{\kappa\alpha}_{aaaaa}\Bigr],
\end{align}
\begin{align*}
z_{aaaab}^{\theta\alpha}=
& W^{\theta\kappa}\times\Bigl[\sigma^{V}[z^{\kappa\alpha}_{a}]^{4}z^{\kappa\alpha}_{b}+\sigma^{IV}(4[z^{\kappa\alpha}_{a}]^{3}z^{\kappa\alpha}_{ab}\\
& +6[z^{\kappa\alpha}_{a}]^{2}z^{\kappa\alpha}_{aa}z^{\kappa\alpha}_{b})+\sigma'''(6[z^{\kappa\alpha}_{a}]^{2}z^{\kappa\alpha}_{aab}+3[z^{\kappa\alpha}_{aa}]^{2}z^{\kappa\alpha}_{b}\\
& +12z^{\kappa\alpha}_{a}z^{\kappa\alpha}_{aa}z^{\kappa\alpha}_{ab}+4z^{\kappa\alpha}_{a}z^{\kappa\alpha}_{b}z^{\kappa\alpha}_{aaa})+\sigma''(4z^{\kappa\alpha}_{a}z^{\kappa\alpha}_{aaab}\\
& +6z^{\kappa\alpha}_{aa}z^{\kappa\alpha}_{aab}+4z^{\kappa\alpha}_{ab}z^{\kappa\alpha}_{aaa}+z^{\kappa\alpha}_{b}z^{\kappa\alpha}_{aaaa})+\sigma'z^{\kappa\alpha}_{aaaab}\Bigr],
\end{align*}
\begin{align*}
z_{aaabb}^{\theta\alpha}= & W^{\theta\kappa}\times\Bigl[\sigma^{V}[z^{\kappa\alpha}_{a}]^{3}[z^{\kappa\alpha}_{b}]^{2}+\sigma^{IV}(3z^{\kappa\alpha}_{a}z^{\kappa\alpha}_{aa}[z^{\kappa\alpha}_{b}]^{2}\\
& +6[z^{\kappa\alpha}_{a}]^{2}z^{\kappa\alpha}_{ab}z^{\kappa\alpha}_{b}+[z^{\kappa\alpha}_{a}]^{3}z^{\kappa\alpha}_{bb})+\sigma'''(3z^{\kappa\alpha}_{a}z^{\kappa\alpha}_{aa}z^{\kappa\alpha}_{bb}\\
& +3[z^{\kappa\alpha}_{a}]^{2}z^{\kappa\alpha}_{abb}+6z^{\kappa\alpha}_{a}z^{\kappa\alpha}_{aab}z^{\kappa\alpha}_{b}+6z^{\kappa\alpha}_{aa}z^{\kappa\alpha}_{ab}z^{\kappa\alpha}_{b}\\
& +z^{\kappa\alpha}_{aaa}[z^{\kappa\alpha}_{b}]^{2}+6z^{\kappa\alpha}_{a}[z^{\kappa\alpha}_{ab}]^{2})+\sigma''(z^{\kappa\alpha}_{bb}z^{\kappa\alpha}_{aaa}\\
& +6z^{\kappa\alpha}_{ab}z^{\kappa\alpha}_{aab}+3z^{\kappa\alpha}_{aa}z^{\kappa\alpha}_{abb}+2z^{\kappa\alpha}_{b}z^{\kappa\alpha}_{aaab}+3z^{\kappa\alpha}_{a}z^{\kappa\alpha}_{aabb})\\
&
+\sigma'z^{\kappa\alpha}_{aaabb}\Bigr].
\end{align*}
In general, a derivative $\partial^{s}$ uses those and
only those derivatives $\partial^{r}$ from the previous layer, for which
$s-r$ has no negative components. For example,
if a cost function would use only $\partial^{5}/\partial a$$^{5}$,
one still needs to calculate $\partial^{4}/\partial a$$^{4}$, $\partial^{3}/\partial a$$^{3}$, $\partial^{2}/\partial a$$^{2}$,
$\partial/\partial a$ and the values themselves for all layers, as seen from formula \eqref{fifth}.
\subsection{Backward pass}
After obtaining the final layer matrices $z^{\omega\alpha}$, $z_{a}^{\omega\alpha}$,
$z_{aa}^{\omega\alpha},\ldots$\textcolor{red}{,} one can calculate a net cost function $E$, usually as the sum of a per-pattern cost function $e$: 
\[
E=\underset{\alpha}{\sum}e(z^{\omega\alpha},z_{a}^{\omega\alpha},z_{aa}^{\omega\alpha},\ldots),
\]
and more importantly, its derivatives with respect to the elements of
all used matrices
\[
\frac{\partial E}{\partial z^{\omega\alpha}},\frac{\partial E}{\partial z_{a}^{\omega\alpha}},\frac{\partial E}{\partial z_{aa}^{\omega\alpha}},\ldots,
\]
which can also be written with a multi-index
\[
\frac{\partial E}{\partial(\partial^{r}z^{\omega\alpha})}.
\]
Those derivatives have indices $\omega$ and $\alpha$, which means they are matrices of the same size as the matrices with respect to which the derivative of $E$ is calculated, in this case $||\omega||\times||\alpha||$. They are propagated
backwards using the following relation (now one moves from $\theta$
to $\kappa$):
\begin{equation}
\label{backformula}
\frac{\partial E}{\partial(\partial^{r}z^{\kappa\alpha})}=\sum_{s}\binom{s}{r}\cdot\partial^{s-r}\sigma'(z^{\kappa\alpha})\cdot\left([W^{\theta\kappa}]^{T}\times \frac{\partial E}{\partial(\partial^{s}z^{\theta\alpha})}\right)
\end{equation}
where sum is taken through all $s$ that were used for
the forward propagation and for which $s-r$
has no negative components. Operator $\partial^{s-r}$ is applied only to $\sigma'(z^{\kappa\alpha})$. Following the notation of \cite{saint2018elementary}, the binomial coefficient is:
\[
\binom{s}{r}=\prod_{i}\frac{s_{i}!}{r_{i}!(s_{i}-r_{i})!}.
\]
Expression \eqref{backformula} is essentially a formula for a derivative upon the transformation of arguments. Consider a simplified scenario with only two derivatives $\partial/\partial a$, $\partial^{2}/\partial a^{2}$. Let $||\alpha||=1$ and $\kappa$ be fixed, $\theta\in[\theta_{1}\dots\theta_{n}]$. $E$ can be considered as the function of values and their derivatives on either of two layers $\theta$ or $\kappa$. For a change of variables
\[E(z^{\theta_{1}},z_{a}^{\theta_{1}},z_{aa}^{\theta_{1}},\dots,z^{\theta_{n}},z_{a}^{\theta_{n}},z_{aa}^{\theta_{n}})\rightarrow E(z^{\kappa},z_{a}^{\kappa},z_{aa}^{\kappa}),\]
the following holds:
\begin{equation}\label{derivzamena}
\frac{\partial E}{\partial z_{a}^{\kappa}}=\sum_{\theta=\theta_{1}}^{\theta_{n}}\frac{\partial E}{\partial z^{\theta}}\frac{\partial z^{\theta}}{\partial z_{a}^{\kappa}}+\frac{\partial E}{\partial z_{a}^{\theta}}\frac{\partial z_{a}^{\theta}}{\partial z_{a}^{\kappa}}+\frac{\partial E}{\partial z_{aa}^{\theta}}\frac{\partial z_{aa}^{\theta}}{\partial z_{a}^{\kappa}}.
\end{equation}
The derivatives of $E$ with respect to $z^{\theta},z_{a}^{\theta}$ and $z_{aa}^\theta$ were calculated on the previous backward step (or initialized at the output layer). The remaining terms can be calculated by differentiating forward pass formulas. Namely, according to \eqref{zeroth} $\partial z^{\theta}/\partial z_{a}^{\kappa}$ is zero, since the values of the next layer $z^{\theta}$ do not depend on the derivatives of the previous one. The term $\partial z_{a}^{\theta}/ \partial z_{a}^{\kappa}$ is obtained by applying $\partial/\partial z_{a}^{\kappa}$ to \eqref{za} (and $\partial z_{aa}^{\theta}/ \partial z_{a}^{\kappa}$ is obtained by applying it to \eqref{mixed} with $a=b$):
\begin{equation}\label{dpodd}
\frac{\partial}{\partial z_{a}^{\kappa}}z_{a}^{\theta}=
\frac{\partial}{\partial z_{a}^{\kappa}}\sum_{\tilde{\kappa}}W^{\theta\tilde{\kappa}}\sigma'(z^{\tilde{\kappa}})z_{a}^{\tilde{\kappa}}=W^{\theta\kappa}\sigma'(z^{\kappa}).
\end{equation}
The matrix multiplication written as a sum over $\tilde{\kappa}$ vanishes as the derivative of a summand is non zero only if $\tilde{\kappa}=\kappa$. Terms like $\sigma'(z^{\kappa})$ are the derivatives of $\partial^{s}\sigma(z^{\kappa})$ (the square brackets in expressions like \eqref{4th}) with respect to one of their terms $\partial^{r}z^{\kappa}$. Their general form is $\binom{s}{r}\partial^{s-r}\sigma'(z^{\kappa})$. It is derived in \cite{avrutskiy2017backpropagation} by the analysis of Fa\`a di Bruno's formula\cite{hardy2006combinatorics}. The sum over $\theta$ in \eqref{derivzamena} can be written as a vector-matrix product (the pattern index $\alpha$ turns it into a matrix-matrix product) so its second term is $\partial E/\partial z_{a}^{\theta}\times \left[W^{\theta\kappa}\cdot\sigma'(z^{\kappa})\right]$. The term $\sigma'(z^{\kappa})$ has no $\theta$ and can be taken out. A transpose and rearrangement to match the dimensions lead to expression \eqref{backformula}.
The gradients of $E$ with respect to weights and thresholds are as follows:
\begin{equation}
\label{de-dw}
\frac{\partial E}{\partial W^{\theta\kappa}}=\sum_{s}\frac{\partial E}{\partial(\partial^{s}z^{\theta\alpha})}\times[\partial^{s}\sigma(z^{\kappa\alpha})]^{T},
\end{equation}
\begin{equation}\label{de-dt}
\frac{\partial E}{\partial t^{k}}=\sum_{\alpha}\frac{\partial E}{\partial z^{\kappa\alpha}}.
\end{equation}
The sum in \eqref{de-dw} runs through all used derivatives. The expression for the gradient of $E$ with respect to the weights $W^{\theta\kappa}$
between layers $\kappa$ and $\theta$ includes matrices like
$z_{a}^{\kappa\alpha}$ which are calculated during the forward pass for layer $\kappa$. They emerge from the term $\partial^{s}\sigma(z^{\kappa\alpha})$. Matrices $\partial E/\partial(\partial^{s} z^{\theta\alpha})$ are obtained for layer $\theta$ during the backward pass. The expression \eqref{de-dw} is also a consequence of a variable change. Let $||\alpha||=1$ and $W^{\theta\kappa}$ be fixed, $\theta\in[\theta_{1}\dots\theta_{n}]$, $s\in[s_{1}\dots s_{m}]$. The change is
\[
E(\partial^{s_{1}}z^{\theta_{1}},\partial^{s_{1}}z^{\theta_{2}},\dots,\partial^{s_{m}}z^{\theta_{n}})\rightarrow E(W^{\theta\kappa}).
\]
The derivative of E is transformed as
\[
\frac{\partial E}{\partial W^{\theta\kappa}}=\sum_{s=s_{1}}^{s_{m}}\sum_{\tilde{\theta}=\theta_{1}}^{\theta_{n}}\frac{\partial E}{\partial(\partial^{s}z^{\tilde{\theta}})}\frac{\partial(\partial^{s}z^{\tilde{\theta}})}{\partial W^{\theta\kappa}}.
\]
According to the forward pass formula \eqref{fpassformula} the latter term is
\[
\frac{\partial(\partial^{s}z^{\tilde{\theta}})}{\partial W^{\theta\kappa}}=\delta_{\theta\tilde{\theta}}\cdot\partial^{s}\sigma(z^{\kappa}).
\]
Upon substitution, the Kronecker delta removes the sum over $\tilde{\theta}$. Pattern index $\alpha$ creates a matrix multiplication and transpose matches the dimensions, which leads to expression \eqref{de-dw}.
\section{Complexity}
\subsection{Forward pass}
Propagation of the derivative $\partial^{s}$ from layer $\kappa$ with $||\kappa||$ neurons to layer $\theta$ with $||\theta||$ neurons, according to \eqref{fpassformula}, requires an element-wise calculation of the term $\partial^{s}\sigma(z^{\kappa\alpha})$ spawned by the chain rule and one matrix multiplication. The number of operations for the latter is $(2||\kappa||-1)\cdot||\theta||\cdot||\alpha||$, and the total number of element-wise operations is $\rho||\kappa||\cdot||\alpha||$, where $\rho$ is the amount of arithmetics per pattern per neuron of layer $\kappa$.

To find $\rho$, that is the number of operations required to calculate the square brackets in  expressions like \eqref{3rd} and \eqref{4th} without taking into account the upper indexes, one may notice that each term is a product of the sigmoid derivative with the order from 1 to $|s|$ and some expression in parentheses. Arithmetics required for the first derivative of the sigmoid \eqref{sigm} can be found by writing it as a polynomial of the $\sigma$ itself:
\[
\sigma'(x)=\frac{e^{-x}}{(1+e^{-x})^{2}}=\frac{1}{1+e^{-x}}-\frac{1}{(1+e^{-x})^{2}}=\sigma-\sigma^{2}.
\]
By differentiating this further, one can obtain all required expressions as polynomials of $\sigma$. Since higher orders are used only together with lower ones, their polynomials can be simplified and expressed in terms of lower order derivatives to decrease the number of operations. Namely, up to the sixth order:
\[
\sigma'=\sigma(1-\sigma) \quad \sigma''=\sigma'(1-2\sigma)
\]
\[
\sigma'''=\sigma'(1-6\sigma') \quad \sigma^{IV}=\sigma''(1-12\sigma')
\]
\[
\sigma^{V}=\sigma'-30(\sigma'')^{2} \quad \sigma^{VI}=\sigma''+(\sigma^{IV}-\sigma'')(5-30\sigma').
\]
The total number of products in the square brackets is equal to the order $|s|$. This, together with the operations required to calculate the derivatives of $\sigma$, but without taking into account the arithmetic of the expressions in parentheses, leads to the following number of extra\footnote{the evaluation of $\sigma$ itself is required for any order. The inclusion of the corresponding number of operations very slightly advantages high-order training and therefore can be ignored in the scope of this paper} operations per pattern per neuron for $|s|$ from 1 through 6: ${3, 8, 13, 18, 23, 30}$. To evaluate the number of operations required to calculate the parentheses, one can notice the following: the parentheses that are multiplied by the $k$\textsuperscript{th} derivative of the sigmoid are the sum of products of $z$ with lower indices which from all possible unique partitions of $s$ into $k$ parts. If some partitions are not unique (if two or more Latin variables are the same), one should multiply the corresponding product by the number of occurrences. Expressions like \eqref{3rd} and \eqref{4th} are the worst-case scenario since all the variables are distinct, therefore, any partition is unique. Since partitions are symmetrical with respect to a permutation of variables, Table \ref{opc} is sufficient for evaluating the number of operations for each derivative used in this paper.
\begin{table}[!t]
	\renewcommand{\arraystretch}{1.3}
	\caption{The number of operations \#op per pattern per neuron for all parentheses of the derivative $\partial^{s}$}
	\label{opc}
	\centering
	\begin{tabular}{c|c||c|c||c|c||c|c}
		\hline
		$s$ & \#op & $s$ & \#op & $s$ & \#op & $s$ & \#op \\
		\hline
		(0) & 0 & (1,1) & 1 & (4) & 11 &(5) & 20\\
		\hline
		(1) & 0 & (3)   & 4 & (3,1) & 19 &(4,1) & 38\\
		\hline
		(2) & 1 & (2,1) & 6 & (2,2) & 22 &(3,2) & 54\\
		\hline
	\end{tabular}
\end{table}
\subsection{Weights gradient}
Expression \eqref{de-dw} for the weights gradient is a sum of matrix products of two terms, one of which was calculated during the backward pass and another is an element-wise part of \eqref{fpassformula} that can be cached during the forward pass. The number of operations for each matrix multiplication is ${||\kappa||\cdot||\theta||\cdot(2||\alpha||-1)}$.
\subsection{Backward pass}
Formula \eqref{backformula} contains a sum for each backpropagated derivative $r$. However, for different $r$, calculations for the summands overlap significantly. Except for the combinatorial coefficient, each summand is an element-wise product of two terms, one of which is a matrix product and another one is very similar to expression $\partial^{s}\sigma(z^{\kappa\alpha})$ encountered in the forward pass \eqref{fpassformula}. The only difference is the presence of $\sigma'$ instead of $\sigma$. One can notice that for $r=0$ multi-index $s-r$ has no negative components for any $s$, therefore, all possible derivatives of $\sigma'(z^{\kappa\alpha})$ have to be calculated for $r=0$ but then can be reused for higher $r$. In fact, all of those terms can be evaluated during the forward pass when the parentheses of expressions like \eqref{3rd} and \eqref{4th} are already calculated and only need to be multiplied by higher derivatives of $\sigma$. Thus, the number of an additional operations per pattern per neuron for $|s-r|$ from 0 through 5 is ${0, 1, 3, 4, 7, 9}$ plus from 2 to 5 operations to increase the maximum order of the derivative of $\sigma$ by one, but only once for the whole bundle of the propagated derivatives. As for the matrix multiplication on the right of \eqref{backformula}, all unique products have to be calculated for $r=0$. For higher $r$, those computationally expensive products should be reused. The only unaccounted part of operations left for $\partial E/\partial(\partial^{r} z^{\kappa\alpha})$ is the multiplications by the binomial coefficients and the summation over all $s$ for which $s-r$ has no negative components. For example, the maximum number of summands is equal to the number of propagated derivatives, which in this paper $\leq21$. One can roughly estimate the ratio between the element-wise and matrix operations as $\frac{\rho}{6\theta}$. For the cases considered in Section III this value never exceeds 0.1, however, caching $z^{\kappa\alpha}$, $\partial^{s}\sigma(z^{\kappa\alpha})$ and $\partial^{s}\sigma'(z^{\kappa\alpha})$ is required. Since this would triple the memory complexity, and the portion of element-wise operations is relatively low, only the neuron activities $z^{\kappa\alpha}$ are cached. Terms $\partial^{s}\sigma(z^{\kappa\alpha})$ and $\partial^{s}\sigma'(z^{\kappa\alpha})$ are calculated on demand. Table \ref{percents} shows the portion of the element-wise operations measured in the percentage of matrix operations provided only the neuron activities $z^{\kappa\alpha}$ are cached. To calculate the equalizing number of epochs for different order training one can simply compare the total amount of matrix operations and then slightly correct it using this table. Note that the exclusion training consists of steps that are the extended training. Table \ref{epochs} generalizes the relative complexity of the extended training for one and two variables derivatives with respect to which are propagated. The first summand is the total number of derivatives. The second summand reflects the element-wise portion of operations. For simplicity, all values were rounded to the nearest integer.
\begin{table}[!t]
	\renewcommand{\arraystretch}{1.5}
	\caption{The amount of element-wise arithmetics for the extended training of order $d$ as the percentage of matrix arithmetics for the examples from Section III}
	\label{percents}
	\centering
	\begin{tabular}{c||c|c|c|c|c|c}
		\hline
		$d$ & 0 & 1 & 2 & 3 & 4 & 5 \\
		\hline
		2D Function & $1\%$ & $2\%$ & $4\%$ & $6\%$ & $10\%$ & $17\%$\\
		\hline
		3D Autoencoder & $2\%$ & $4\%$ & $10\%$ & $15\%$ & $22\%$ & $30\%$\\
		\hline
		Poisson's equation & $7\%$ & $12\%$ & $19\%$ & $31\%$ & - & -\\
		\hline
	\end{tabular}
\end{table}
\begin{table}[!t]
	\renewcommand{\arraystretch}{1.5}
	\caption{The relative complexity of one epoch of the extended training of order $d$ for a network with layer configuration $n^{*},n,n,n,n,n^{*}$ for one and two variables for differentiation}
	\label{epochs}
	\centering
	\begin{tabular}{c||c|c|c|c|c|c}
		\hline
		order & 0 & 1 & 2 & 3 & 4 & 5 \\
		\hline
		$a$ & $1$ & $2+\frac{2}{5n}$ & $3+\frac{6}{n}$ & $4+\frac{15}{n}$ & $5+\frac{28}{n}$ & $6+\frac{48}{n}$\\
		\hline
		$a,b$ & $1$ & $3+\frac{2}{5n}$ & $6+\frac{13}{n}$ & $10+\frac{45}{n}$ & $15+\frac{120}{n}$ & $21+\frac{280}{n}$\\
		\hline
	\end{tabular}
\end{table}
\subsection{Hardware efficiency.}
Some technical details are required to get the maximum hardware efficiency. This study uses CUDA with cuBLAS\cite{nvidia2008cublas}, that is, a GPU-accelerated implementation of the standard basic linear algebra subroutines (BLAS). Its function cublasSgemm is used to compute matrix multiplications. The rest of the operations are element-wise and can be implemented by a C-like code that describes the computations on one element which are then parallelized automatically across the whole matrix. The code should be written in such a way that delays between the calls of cublasSgemm and other functions are minimal, and matrices which they operate on are as large as possible. The heights of matrices are obviously fixed by the network's configuration, so they can only be as long as possible, i.e. it is preferably to process all input patterns in one batch. An efficient method to make matrices longer is to use pattern index $\alpha$ to stack the matrices of different derivatives together. In the forward pass formula \eqref{fpassformula}, the matrix multiplication $W^{\theta\kappa}\times\left[\partial^{s}\sigma(z^{\kappa\alpha})\right]$ is supposed to be called for each $s$ separately. But if one first evaluates all (let us say $n$) different terms $\partial^{s}\sigma(z^{\kappa\alpha})$ and stores them in the memory as a single matrix with a new index $\widetilde{\alpha}\in[1\dots n||\alpha||]$ then cublasSgemm can be called only once and will result in a similarly stacked $\partial^{s}z^{\theta\widetilde{\alpha}}$ matrix. This requires the column-wise storage of matrices and an explicit memory allocation but, for example, made the calculations of the 5\textsuperscript{th} order for 2D function approximation few times more efficient. Backward pass formula \eqref{backformula} allows a similar enhancement. 

As for the delay between the functions calls, two options are available: either memory for the matrices of all derivatives of all layers is permanently allocated on the GPU or intermediate data saves/loads occur in the background while some matrices are propagated from one layer to another. The first option is suitable for systems with enough GPU memory. The second option can be implemented in any system with fast enough communications between the device and the host. If the time for a matrix $||\alpha||\times||\kappa||$ to load is less than the time required to multiply it from the right by a matrix $||\kappa||\times||\theta||$ then one can hide all memory operations behind the computations.

A trick to reduce the memory complexity of formula \eqref{backformula} is worth mentioning. Instead of
calculating matrix products $[W^{\theta\kappa}]^{T}\times$$\partial E/\partial(\partial^{s}z^{\theta\alpha})$
and storing them in the cache one can put them in the memory
dedicated for $\partial E/\partial(\partial^{r}z^{\kappa\alpha})$
with $r=s$ and then calculate $\partial E/\partial(\partial^{r}z^{\kappa\alpha})$
in a proper order. Namely, as soon as the evaluation is started for any $r$, one has to multiply a matrix product residing in that memory by $\sigma'(z^{\kappa\alpha})$, thus, it
cannot be reused. To tackle this issue, one should
start with $r=(0,0,0,0)$ and, thus, spoil $s=(0,0,0,0)$,
which is not used for any other $r$ since the components of $s-r$ can not be negative. Then one can pick any first order derivative, for
example, $r=(1,0,0,0)$ and spoil $s=(1,0,0,0)$,
which is not a problem, since the only other case when $(1,0,0,0)-r$ has no negative components is when $r=(0,0,0,0)$, but that term has already been calculated. All the first order derivatives are calculated, then all the second order ones and so on, and no conflicts are encountered.
\subsection{Performance test}
CUDA C code was written using the proposed suggestions. It was tested on fully connected perceptrons with 7 layers of the same width $n$. RProp with the regular cost was run for 1000 epochs with 2048 input patterns processed in one batch. It is compared against Keras 2.0.8 with backends theano\cite{bergstra2011theano} 0.9.0 and TensorFlow\cite{abadi2016tensorflow} 1.3.0, default settings are used. The system is Deep Learning AMI for Amazon Linux, version 3.3 run on EC2 p2.xlarge instance with one GK210 core of Tesla K80 available. The driver version is 375.66, CUDA version is 8.0. The results are gathered in Table \ref{ptab}. This paper uses neural networks with $n$ equal to 64 and 128. Provided TensorFlow can scale its performance for cases when many derivatives are being propagated, the gain is around 300\%. Even for the regular training where standard neural network libraries should be quite efficient, the proposed code is about 3 times faster for the networks used in this paper. For cases where many derivatives are to be calculated, naive implementations of automatic differentiation would probably be much slower.
\begin{table}[!t]
	\renewcommand{\arraystretch}{1.3}
	\caption{Running times for the regular training}
	\label{ptab}
	\centering
	\begin{tabular}{c||c||c||c}
		\hline
		\bfseries $n$ & \bfseries Theano & \bfseries TensorFlow & \bfseries CUDA C \\
		\hline\hline
		128 & 17.4 & 10 & 3.1\\
		\hline
		256 & 19.9 & 11.8 & 5.7\\
		\hline
		512 & 27.6 & 22.2 & 14.9\\
		\hline
	\end{tabular}
\end{table}

\section{Conclusion}
A training process that enhances the approximation abilities of fully connected feedforward neural networks was presented. It is based on calculating extra derivatives of the network and comparing them with the target ones to evaluate the weights gradient. It was demonstrated to work well for low-dimensional cases. Using derivatives up to the 5\textsuperscript{th} order, the precision of approximation for 2D analytical function was increased 1000 times. Among all derivatives, the first and the second contributed the most to the relative increase of accuracy. Computational costs per pattern increase significantly (see table \ref{epochs}), however, it seems that there are no conditions under which the conventional training could catch up with the proposed one provided a network capacity is sufficient. High-order training was found to be more demanding in this regard. For 2D approximation lowering the number of neurons in all hidden layers from 128 to 64 and then to 32 increased the root mean square of the deviation of values for the 5\textsuperscript{th} order extended training from $2.2\cdot10^{-6}$ to $2.4\cdot10^{-5}$ and then to $2.4\cdot10^{-4}$, for the 2\textsuperscript{nd} order from $3.7\cdot10^{-5}$ to $1\cdot10^{-4}$ and then to $5\cdot10^{-4}$ and for the regular training from $2.3\cdot10^{-3}$ to $3.0\cdot10^{-3}$ and then to $5.3\cdot10^{-3}$. Increasing the number of neurons higher than 128 was not beneficial for any order.

For real neural network applications like classification one can imagine hard times calculating high-order derivatives. From this point of view, solving partial differential equations can benefit much more from the proposed enhancements, as all information about extra derivatives can be obtained via simple differentiation of the equation itself. In the presented example of solving partial differential equation, the precision of the regular method was quite acceptable, however, it required a grid with three times smaller spacing. Even though the extended training achieved 13 times smaller error for the same computational cost, the increase of a spacing $\lambda$ might be more important. As the number $N$ of dimensions increases, the grid size grows as $(1/\lambda)^{N}$. If a similar increase of the grid spacing persists for high-dimensional partial differential equations, the extended training could be even more advantageous.


%

\appendices

\section*{Acknowledgment}
Author owes a great debt of gratitude to his scientific advisor E.A.~Dorotheyev. Special thanks for invaluable support are due to Y.N.~Sviridenko, A.M.~Gaifullin, I.A.~Avrutskaya and I.V.~Avrutskiy without whom this work would be impossible. Sincere gratitude is extended to Neural Networks and Deep Learning lab, MIPT and M.S.~Burtsev personally.




%



\bibliographystyle{IEEEtran}
\bibliography{bibfile}

%








\end{document}